# Shape Completion using 3D-Encoder-Predictor CNNs and Shape Synthesis

Angela Dai[1]   Charles Ruizhongtai Qi[1]   Matthias Nießner[1,2]

[1]Stanford University   [2]Technical University of Munich

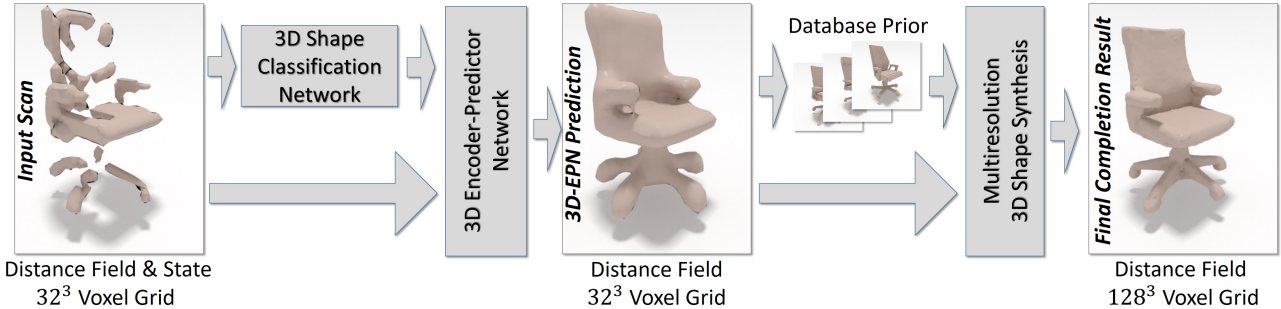

Our method completes a partial 3D scan using a 3D Encoder-Predictor network that leverages semantic features from a 3D classification network. The predictions are correlated with a shape database, which we use in a multi-resolution 3D shape synthesis step. We obtain completed high-resolution meshes that are inferred from partial, low-resolution input scans.

## Abstract

*We introduce a data-driven approach to complete partial 3D shapes through a combination of volumetric deep neural networks and 3D shape synthesis. From a partially-scanned input shape, our method first infers a low-resolution – but complete – output. To this end, we introduce a 3D-Encoder-Predictor Network (3D-EPN) which is composed of 3D convolutional layers. The network is trained to predict and fill in missing data, and operates on an implicit surface representation that encodes both known and unknown space. This allows us to predict global structure in unknown areas at high accuracy. We then correlate these intermediary results with 3D geometry from a shape database at test time. In a final pass, we propose a patch-based 3D shape synthesis method that imposes the 3D geometry from these retrieved shapes as constraints on the coarsely-completed mesh. This synthesis process enables us to reconstruct fine-scale detail and generate high-resolution output while respecting the global mesh structure obtained by the 3D-EPN. Although our 3D-EPN outperforms state-of-the-art completion method, the main contribution in our work lies in the* combination *of a data-driven shape predictor and analytic 3D shape synthesis. In our results, we show extensive evaluations on a newly-introduced shape completion benchmark for both real-world and synthetic data.*

## 1. Introduction

Since the introduction of commodity range sensors such as the Microsoft Kinect, RGB-D scanning has gained a huge momentum in both offline and real-time contexts [28, 3, 30, 45, 4, 8]. While state-of-the-art reconstruction results from commodity RGB-D sensors are visually appealing, they are far from usable in practical computer graphics applications since they do not match the high quality of artist-modeled 3D graphics content. One of the biggest challenges in this context is that obtained 3D scans suffer from occlusions, thus resulting in incomplete 3D models. In practice, it is physically infeasible to ensure that all surface points are covered in a scanning session, for instance due to the physical sensor restrictions (e.g., scan behind a shelf, or obtain the fine structure of chair model).

Even when reducing the scope to isolated objects, the problem remains challenging. While traditional methods can fill in small holes via plane fitting, Laplacian hole filling [41, 27, 50], or Poisson Surface reconstruction [16, 17], completing high-level structures, such as chair legs or airplane wings, is impractical with these geometry processing algorithms.

One possible avenue is based on recent advances in machine learning, which suggests that data-driven approaches may be suitable for this task. For instance, assuming a partial 3D scan, one would want to complete the 3D shape geometry based on a previously learned prior.

In this paper, we explore the feasibility of directly applying deep learning as a strategy to predict missing structures from partially-scanned input. More specifically, we propose 3D-Encoder-Predictor Networks (3D-EPN) that are based on volumetric convolutional neural nets (CNNs). Here, our aim is to train a network that encodes and generalizes geometric structures, and learns a mapping from partial scans to complete shapes, both of which are represented as implicit distance field functions. One of the insights of the 3D-EPN is that it leverages semantics from a classification network. More specifically, we use the probability class vector of a 3D-CNN as input to the latent space of the 3D-EPN. In order to provide supervised training data, realistic ground truth scanning patterns are generated from virtually scanned 3D CAD models.

---
*This research is funded by Google Tango.

In our results, we show that 3D-EPNs can successfully infer global structure; however, it remains challenging to predict local geometric detail. In addition, increasing the output resolution comes with significant compute costs and makes the optimization of the training process much more difficult due to the cubic behavior of 3D space. However, we argue that it may be sufficient to predict only coarse (potentially blurry) 3D geometry without fine-scale detail if we can correlate these low-resolution predictions with high-resolution 3D geometric signal from a shape database. As the second technical component, we learn this correlation by searching for similar shapes, and we provide an iterative optimization strategy to incorporate low-level geometric priors from the database in a shape synthesis process.

Hence, we propose a 3D shape synthesis procedure to obtain local geometric detail. Thus, output is synthesized at a much higher resolution than efficiently tractable with 3D deep networks. We first learn a correlation between the predictions of our 3D-EPNs and the CAD models in the database. To this end, we utilize the feature learning capabilities of volumetric CNNs that provide an embedding where 3D-EPNs results are close to geometrically similar CAD models in the database. We learn this embedding as a byproduct of a discriminative classification task. In an iterative optimization procedure, we then synthesize high-resolution output from the 3D-EPN predictions and the database prior.

Overall, we propose an end-to-end mesh completion method that completes partial 3D scans even in very challenging scenarios. We show compelling results on this very challenging problem on both synthetic and real-world scanning data. In addition, we favorably compare against state-of-the-art methods both qualitatively and quantitatively.

In summary, our contributions are

- a 3D-Encoder-Predictor Network that completes partially-scanned 3D models while using semantic context from a shape classification network.
- a 3D mesh synthesis procedure to obtain high-resolution output and local geometric detail.
- an end-to-end completion method that combines these

two ideas, where the first step is to run the 3D ConvNet regressor, and the second step is an iterative optimization for 3D shape synthesis.

## 2. Previous Work

**Shape Completion** Shape completion has a long history in geometry processing, and is often used in the context of cleaning up broken 3D CAD models. In particular, filling in small holes has received much attention; for instance, one could fit in local surface primitives, such as planes or quadrics, or address the problem with a continuous energy minimization; e.g., with Laplacian smoothing [41, 27, 50]. Poisson surface reconstruction can be seen as part of this category [16, 17]; it defines an indicator function on a (potentially hierarchical) voxel grid which is solved via the Poisson equation.

Another direction for completing shapes is detecting structures and regularities in 3D shapes. For instance, many works detect symmetries in meshes or point clouds, and use them to fill in missing data [44, 25, 32, 39, 42]. Although these methods show impressive results, using pre-defined regularities fundamentally limits the shape space to the hand-crafted design.

Much research leverages strong data-base priors. Sung et al. [43] combine this idea with the detection of symmetries and operate on part-based model obtained from the database. Another idea is to find identical CAD models in a shape database for a given partial input shape and align it with the scan [26, 36, 18, 23, 38]. Given the advances in geometric feature matching, it is possible to find these connections; however, these approaches rely on the assumption that the database includes identical (or at least very similar) shapes; thus, they cannot generalize easily to new shapes. To address this shortcoming, one possibility is to first retrieve similar shapes from a database, and then modify the retrieval results such that they better match and explain the partially-scanned input. This way, the retrieved models do not have to exactly match, and it is possible to cover a wider range of objects even with a relatively small database. For instance Pauly et al. [31] complete 3D scans by first retrieving candidate models from a database, then perform a non-rigid mesh alignment and blend the results with the input scan. The same strategy can be applied directly on range images. Rock et al. [34] retrieve similar depth images which they deform in order to predict missing voxels; as a final step they perform Poisson surface reconstruction obtain the resulting mesh. Li et al. [22] use single RGB-D images as input and run a similar pipeline, where they first find and deform nearest models form a database. As a final step they perform a shape synthesis step, which is similar than ours. While the idea of non-rigidly deforming models from a database improves shape coverage, the major limitation is still that global structure cannot be easily generalized

(e.g., high-level structural changes). In our method we also rely on geometric signal from database lookups at test time; however, one of the key insights is that we only take this information into account to synthesize local detail rather than global structure.

In order to generalize to arbitrary new shapes, fully data-driven methods trained with modern machine learning techniques are a promising direction. One of the first methods in this space is Voxlets [11]. They train a random decision forests that predict unknown voxel neighborhoods; the final mesh is generated with a weighted average of the predicted results and by running marching cubes. 3D ShapeNets [2] is probably most related to our 3D Encoder-Predictor network. They also use convolutional neural networks – specifically a deep belief network – to obtain a generative model for a given shape database. This allows them to predict multiple solutions conditioned on partial input; however, as we demonstrate in our results, this strategy is significantly less efficient than directly training an end-to-end predictor as our 3D-EPN does. Nguyen et al. [29] build on this work and apply it to repairing meshes; they use the input of 3D ShapeNets and compute a distance transform on which they apply a Markov Random Field.

**Related Deep Learning Works** With recent advances in machine learning and the availability of 3D shape databases [48, 2], research has started to look at deep learning approaches on 3D data. Wu et al. [48] were among the first that proposed the use of 3D-CNNs for both classification and generative tasks (see above). They use a volumetric representation in their deep belief network that is trained on their own database; although the training is in 3D most of their input is from single range images. Since then, different versions of 3D-CNN architectures have been proposed in order to improve classification accuracy [24, 33], obtain object proposals [40], match local 3D geometry [49], or denoise shapes [37]. While the denoising approach of Sharma et al. [37] can be used towards shape completion, they focus on random noise patterns rather than partial range scans. In this work, we leverage the advances in 3D deep learning and apply a 3D convolutional net for the shape completion task. While previous works focus more on discriminative tasks on shape classification, our network regresses missing data conditioned on the partial scan input.

Recently deep learning has also explored models for generative tasks; for instance, with generative adversarial networks (GANs) [12, 21, 20, 47]. Here, an image (or potentially a 3D shape) is generated from scratch by only taking a random, latent vector as input. This is related and highly interesting direction (in particular, for modeling applications); however, it is well known that current generative models face resolution limits and are usually very hard to train. In our work, we take a more direct path to train a convolutional network to directly predict the missing part of a shape with a follow up shape synthesis module.

## 3. Method Overview

The goal of our method is to take a partial 3D scan of an object as input, and predict a completed 3D shape as output. To achieve this task, we represent each model in a 3D voxel grid. Instead of using just an occupancy grid, we compute the distance transform for all train and test data. For generating ground truth train pairs, we virtually scan objects from the ShapeNet dataset [2] for input, and use a 3D digital differential analyzer [1] to obtain the complete distance field; see Sec. 4.

Once we have generated the training set, we feed the training pairs into a deep neural network which directly operates on the 3D representation. The networks loosely follows idea of autoencoders, similar to Dosovitskiy [10]; however, in our case, we *filter* a volumetric representation, on which we also define the loss function; see Sec. 5. Unlike traditional autoencoder networks that reconstruct the original input and learn an efficient encoding, we aim to fill in missing data from partial input scans. In our case, the network learns a correlation of partial and complete models at training time, which at test time regresses a completed model with constraints given by known surfaces or free space information. On a high level, the goal is to map all partial scans into a shared, embedded space which we correlate with the complete models. We design the training process such that we learn this mapping, as well as the reconstruction from it, even under largely missing data. Here, the main objective is the ability to reconstruct a complete mesh from the latent space while respecting the constraints of known data points.

The main challenge of this process is generating new information – i.e., filling in the missing data from unseen views – by generalizing geometric structures. The network needs to encode general rules of 3D model design, and generalize across different shape instances. To this end, we train the network under input from a shape classification network in oder to respect and leverage semantic information of the shape's geometry. Specifically, we input the probability class vector of a 3D-CNN classification output into the latent space of the 3D-EPN. Another important challenge on 3D shape completion is the high dimensionality; one of the insights here is that we use a (mostly) continuous distance field representation over an occupancy grid; this allows us to formulate a well-suited loss function for this specific task.

Since regressing high-dimensional output with deep networks is challenging for high-resolutions – particularly in 3D space –, we expect the 3D-EPN to operate on a relatively low voxel resolution (e.g., $32^3$ voxel volumes). Although it lacks fine geometric detail, it facilitates the prediction of (missing) global structures of partially-scanned objects (e.g., chair legs, airplane wings, etc.). At test time, we use the ShapeNet database [2] as a powerful geometric prior, where we retrieve high-resolution geometry that respects the high-level structure of the previously obtained predictions. We establish correlations between the low-resolution 3D-EPN output and the database geometry by learning a geometry lookup with volumetric features. Here, we utilize the feature learning of volumetric convolutional networks with a modified version of Qi et al. [33] whose learned features are the byproduct of a supervised classification network. For a given 3D-EPN prediction, we then run the 3D feature extraction and look up the three nearest shape neighbors in the database which are most similar regarding the underlying geometric structure.

As a final step of our completion pipeline, we correlate the coarse geometric predictions from the 3D-EPN output with the retrieved shape models. We then synthesize higher resolution detail by using the retrieved shape models to find similar volumetric patches to those in our prediction, and use these to iteratively optimize for a refined prediction, hierarchically synthesizing to a $128^3$ high-resolution distance field. This effectively transfers-high resolution detail from complete, synthetic shapes to the prediction while maintaining its intrinsic shape characteristics. From this implicit surface representation, we then extract the final mesh from the isosurface.

## 4. Training Data Generation

For training data generation, we use the ShapeNet model database [2], and we simultaneously train on a subset of 8 categories (see Sec. 8) and a total of 25590 object instances (the test set is composed of 5384 models). In the training process, we generate partial reconstructions by virtually scanning the 3D model. Here, we generate depth maps from random views around a given model with our custom virtual DirectX renderer. The obtained depth maps store range values in normalized device coordinates. We backproject these to metric space (in $m$) by using Kinect intrinsics. The extrinsic camera parameters define the rigid transformation matrices which provide alignment for all generated views. All views are integrated into a shared volumetric grid using the volumetric fusion approach by Curless and Levoy [6], where the voxel grid's extent is defined by the model bounding box. Note that the ground truth poses are given by the virtual camera parameters used for rendering and the models are aligned with respect to the voxel grid. As a result, we obtain a truncated signed distance field (TSDF) for a given (virtual) scanning trajectory. This representation also encodes known free space; i.e., all voxels in front of an observed surface point are known to be empty. The sign of the distance field encodes this: a positive sign is known-empty space, zero is on the surface, and a negative sign indicates unknown values. This additional information is crucial for very partial views; see Fig. 2. For training the 3D-EPN, we separate our the sign value from the absolute distance values, and feed them into the network in separate channels; see Sec. 5.

For each model, we generate a set of trajectories with different levels of partialness/completeness in order to reflect real-world scanning with a hand-held commodity RGB-D sensor. These partial scans form the training input. The ground truth counterpart is generated using a distance field transform based on a 3D scanline method [1]; here, we obtain a *perfect* (unsigned) distance field (DF). We choose to represent the ground truth as an unsigned distance field because it is non-trivial to robustly retrieve the sign bit from arbitrary 3D CAD models (some are closed, some not, etc.). In our training tasks, we use six different partial trajectories per model. This serves as data augmentation strategy, and results in a total of $153,540$ training samples of our 3D-EPN.

Within the context of this paper, we generate training pairs of TSDF and DF at resolutions of $32^3$. The final resolution of our completion process is an implicit distance field representation stored in volumes of $128^3$ voxels after we apply the shape synthesis step; see Sec. 7.

## 5. 3D Encoder-Predictor Network (3D-EPN) for Shape Completion

We propose a 3D deep network that consumes a partial scan obtain from volumetric fusion [6], and predicts the distance field values for the missing voxels. Both our input and output are represented as volumetric grids with two channels representing the input TSDF; the first channel encodes the distance field and the second known/unknown space; see Sec. 4. Note that the binary known/unknown channel encodes a significant amount of knowledge as well, it will let the network know what missing areas it should focus on.

Our network is composed of two parts and it is visualized in Fig. 1. The first part is a 3D encoder, which compresses the input partial scan. The compressed stream is then concatenated with the semantic class predictions of a 3D-CNN shape classifier into a hidden space volume; the input partial scan is compressed through a series of 3D convolutional layers, followed by two fully-connected layers which embed the scan and its semantic information into the latent space. This encoder helps the network summarize global context from the input scan – both the observed distance values, known empty space, and class prediction. The second part is a predictor network that uses 3D up-convolutions to grow the hidden volume into a $32^3$ full size output of estimated distance field values. Based on the global context summarized by the encoder network, the predictor net is able to infer missing values. In addition, we add skip connections – similar to a U-net architecture [35] – between the corresponding encoder and predictor layers, visualized at the bottom of Fig. 1. The data from these connections is then concatenated with the intermediary output of the up-convolutions, thus doubling the feature map size. This way, we ensure propagation of local structure of the input data and make sure it is preserved in the generated output predictions.

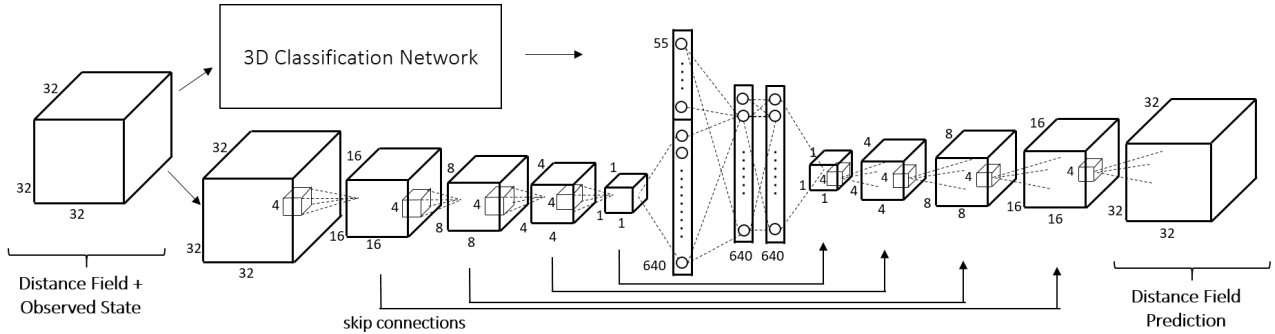

Figure 1: Network architecture of our 3D Encoder-Predictor Network.

We use ReLU and batch normalization for all the layers (except the last one) in the network. We use a masked L1 loss that computes the difference of ground truth distance field and predicted ones. Only the error in the unknown regions is counted; the known occupied and known empty voxels are masked out and enforced to match up the input. We use the ADAM optimizer [19] with $0.001$ learning rate and momentum $0.9$. The learning rate is decayed by half every 20 epochs. For $153,540$ training samples, it takes $\approx 3$ days to train the model to convergence (about half as long without the skip connections).

## 6. Shape Prior Correlation

Our 3D Encoder-Predictor Network predicts a $32^3$ distance field from partially-scanned input. To generate high resolution detail from this coarse prediction, we correlate these predictions with 3D CAD models from a shape database. To this end, we learn a shape feature descriptor with a 3D-CNN using a modified architecture of Qi et al. [33]. The network is trained as a classification task on all 55 classes of ShapeNet, which provides a powerful learned feature descriptor. Since the descriptor is obtained by training on object categorization, it also defines an embedding of shape similarities. This allows us to perform shape similarity queries between the 3D-EPN predictions and the CAD model database.

For the shape completion, we assume that we have access to all training meshes of ShapeNet at their full resolution; i.e., we use the shape database as geometric prior rather than encoding all fine-scale detail in a 3D deep net. Based on the learned feature vector, we retrieve the three closest models from the database that are most similar to the 3D-EPN output; this is a k-nearest-neighbor query based on geometric similarity. In all of our experiments, we exclude the 5397 models from the test benchmark; hence, ground truth models cannot be retrieved and are not part of the feature learning. Although in real-world scanning applications it is a valid scenario that physical and virtual objects are identical (e.g., IKEA furniture), we did not further explore this within the context of this paper since our aim is to generalize to previously unseen shapes.

## 7. Shape Synthesis and Mesh Generation

In this section, we describe how we synthesize the final high-resolution output and generate local geometric detail. Here, the input is the prediction of the 3D-EPN, as described in Sec. 5, as well as the nearest shape neighbors obtained from the shape prior correlation as described in Sec. 6. We then run an iterative shape synthesis process that copy-pastes voxels from the nearest shape neighbors to construct a high-resolution output from the low-resolution predictions.

Similar to Hertzmann et al. [13], our volumetric synthesis searches for similar volumetric patches in the set of $k$ nearest shape neighbors to refine the voxel predictions from the 3D-EPN. Let $P$ be the low resolution output of the 3D-EPN, of dimension $d_0 \times d_0 \times d_0$ (we have $d_0 = 32$). Multi-scale pyramids are computed for the $k$ shape neighbors, with each level $l$ containing the distance field transform of the shape at dimension $2^l d_0$. We synthesize from coarse to fine resolution, initializing with the coarse prediction $P$ and computing a multi-scale representation of $P'$. For every level, volumetric patch features are computed for each voxel of the neighbors $\{N_1^l, ..., N_k^l\}$. To synthesize level $l$ of $P'$, we compute the volumetric patch feature for each voxel $v$ and use an approximate nearest neighbor search [14] to find the most similar voxel $w$ of the neighbors, and update the value of $P'(v)$ with that of $N_x^l(w)$.

The feature for a voxel $v$ at level $l$ is computed from the distance field values of the $5 \times 5 \times 5$ neighborhood of $v$ at level $l$ as well as the values in the corresponding $3 \times 3 \times 3$ neighborhood at level $l-1$. We concatenate these together and perform a PCA projection over the features of $\{N_1^l, ..., N_k^l\}$ to dimension 100 to accelerate the search. Additionally, we only consider features for voxels whose neighborhoods contain at least one voxel on the isosurface of the distance field; i.e., we only synthesize voxels near the surface.

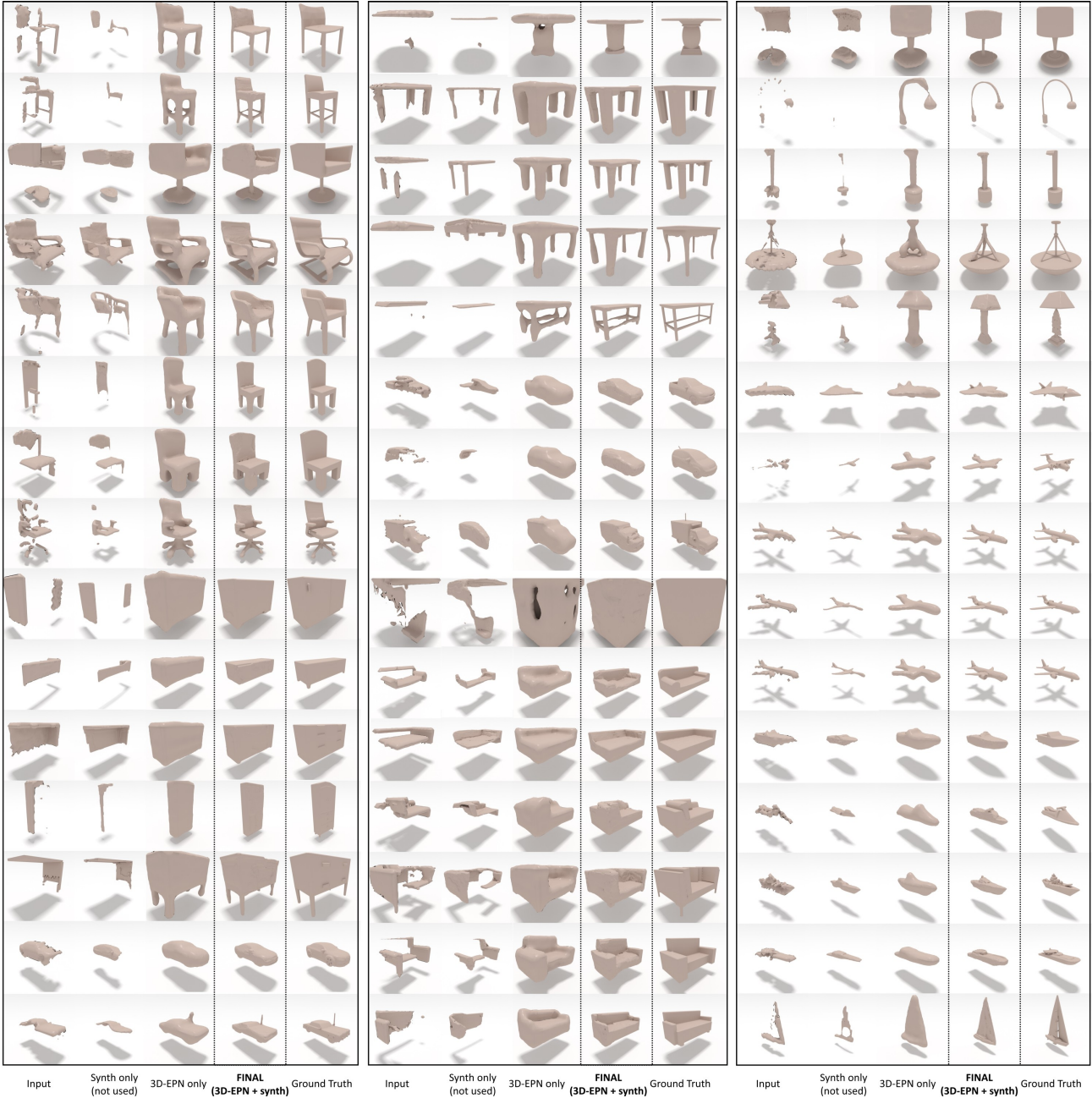

Figure 2: Example shape completions with our method (note that our approaches operates on all shape types using the same trained models). We break out the results of separate steps. For instance, this shows what happens when the shape synthesis step was directly applied to the input; here, we miss global structures.

Thus, we can hierarchically synthesize to an output resolution of $128^3$ voxels, where every voxel contains a distance value. The final step after the mesh synthesis process, is the mesh extraction from the implicit distance field function using Matlab's *isosurface* function.

## 8. Results

Across all experiments, we train the 3D-CNN classifier network, the 3D-EPN, and the 3D retrieval network on the same train/test split for ShapeNet [2], with the 3D-EPN trained on a subset of eight classes: namely, airplanes, tables, cars, chairs, sofas, dressers, lamps, and boats. Quantitative evaluations are obtained for a test set of 1200 models. When a distance field representation is available, we extract the isosurface using Matlab's *isosurface* function. However, some baselines directly predict meshes; in these cases, we use those for rendering and evaluation.

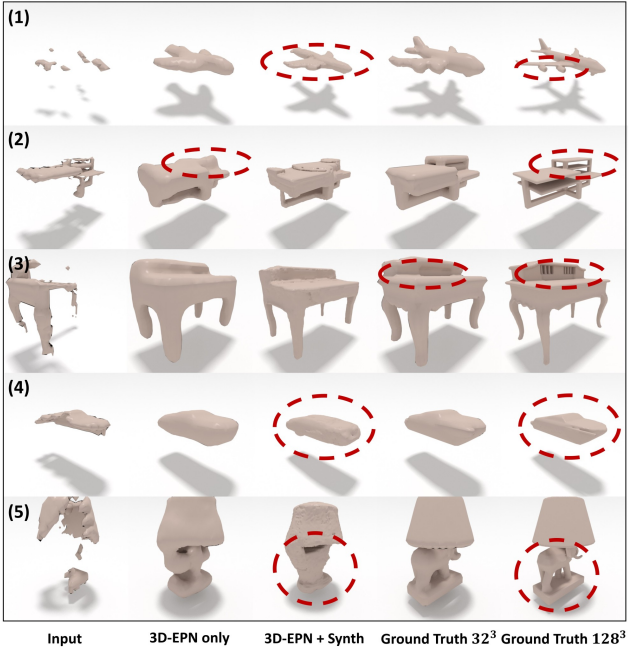

Figure 3: Limitations: (1) in cases of extreme partial input, we fail to infer some structures; (2),(3) fine-scale structures are often missing in the low-resolution ground truth ($32^3$ volume is used as a target for the 3D-EPN); (4) in some cases, semantic predictions are wrong (here, a boat is turned into a car); (5) some shapes are just strange (a lamp with an elephant).

Fig. 2 shows a variety of the test examples of our approach. In each column, we first show the partial input, then we show results where only the 3D synthesis is used. In this experiment, we see that the synthesis alone is unable to complete missing geometric structure (this is not an actual result). Next, we show the results of the 3D-EPN without the synthesis; here, we see that structure is completed but locally the geometry has a low resolution. This is addressed by the combination of 3D-EPN and 3D synthesis, which provides both global structure and local accuracy. In the right of each column, we show the ground truth.

In Fig. 4, we compare against state-of-the-art shape completion methods. Poisson surface reconstruction [16, 17] is mostly used to obtain complete surfaces on dense point clouds, but it cannot infer missing structures. ShapeRecon [34] performs slightly better, but overall, it is heavily dependent on finding good nearest neighbors; the available implementation was also trained only on a subset of classes. 3D ShapeNets [48] is most similar to our method, but it is a fully generative model, which in practice hurts performance since it addresses a more general task. A quantitative evaluation on the same dataset is shown in Tab. 1. Overall, our 3D-EPN performs best, and it efficiently leverages the 3D-CNN class vector input. Our final result at is obtained with the combination of the 3D-EPN and 3D shape synthesis, which outputs a distance field at $128^3$ voxels.

| Method | $\ell_1$-Err ($32^3$) | $\ell_1$-Err ($128^3$) |
|---|---|---|
| Poisson [16, 17] | 1.90 | 8.46 |
| ShapeRecon [34] | 0.97 | 4.63 |
| 3D ShapeNets [48] | 0.91 | 3.70** |
| Ours (synth-only) | 1.20 | 6.92 |
| Ours (3D-EPN) | 0.51 | 2.63** |
| Ours (3D-EPN-class) | 0.48 | 2.48** |
| Ours (3D-EPN-unet) | 0.38 | 2.29** |
| Ours (3D-EPN-unet-class) | 0.37 | 2.29** |
| Ours (3D-EPN + synth) | - | 2.33 |
| Ours (3D-EPN-class + synth) | - | 2.16 |
| Ours (3D-EPN-unet + synth) | - | 1.91 |
| **Ours (final)** 3D-EPN-unet-class + synth | - | **1.89** |

Table 1: Quantitative shape completion results on synthetic ground truth data. We measure the $\ell_1$ error of the unknown regions against the ground truth distance field (in voxel space, up to truncation distance of 2.5 voxels). **predictions at $128^3$ are computed by upsampling the low-resolution output of the networks.

| | 3D-CNN /w Partial Train | 3D-EPN + 3D-CNN /w Complete Train |
|---|---|---|
| Classification | 90.9% | **92.6%** |
| Shape Retrieval | 90.3% | **95.4%** |

Table 2: Effect of 3D-EPN predictions on classification and shape retrieval tasks. We train a 3D-CNN classification network [33] on partial (left) and complete (right) ShapeNet models. The retrieval accuracy is computed from the classes of the top 3 retrieved neighbors. Performance improves significantly when we use the 3D-EPN predictions as an intermediary result. Note that the test task is the same for both cases since they use the same test input.

In Tab. 2, we address the question whether it is possible to use the 3D-EPN to improve accuracy on classification and retrieval tasks. For a given partial scan, there are two options to perform classification. In the first variant, we train the 3D-CNN of Qi et al. [33] on partial input to reflect the occlusion patterns of the test data. In the second variant, we first run our 3D-EPN and obtain a completed $32^3$ output; we use this result as input to the 3D-CNN which is now trained on complete shapes. In both cases, the exact same partial test inputs are used; however, with the intermediate completion step, performance for both classification and shape retrieval increases significantly.

Limitations are shown in Fig. 3. The most important limitation is the rather low resolution of the 3D-EPN. While it successfully predicts global structure, it fails to infer smaller components. This is particularly noticeable when geometric detail is below the size of a voxel; note that the 3D-EPN ground truth training pairs are both at a resolution of $32^2$ voxels. Another limitation is extreme partial input where not enough context is given to infer a plausible completion. However, note that in addition to occupied surface voxels, the test input's *signed* distance field also encodes known-empty space. This is crucial in these cases. A general problem is the availability of 3D training data. With the models from ShapeNet [2], we can cover some variety; however, it is certainly not enough to reflect all geometries of real-world scenes. For further results and evaluation, we refer to the appendix. We show completion results on Kinect scans and evaluate the importance of the signed distance field representation over other representations, such as occupancy or ternary-state voxel grids.

## 9. Conclusion and Future Work

We have presented an efficient method to complete partially-scanned input shapes by combining a new 3D deep learning architecture with a 3D shape synthesis technique. Our results show that we significantly outperform current state-of-the-art methods in all experiments, and we believe that a combination of deep learning for inferring global structure and traditional synthesis for local improvements is a promising direction.

An interesting future direction could be to combine purely generative models with conditioned input, such as GANs [12]. However, these networks are challenging to train, in particular for higher resolutions in 3D space. Another possible avenue is the incorporation of RGB information; for instance, one could enforce shading constraints to obtain fine-scale detail by borrowing ideas from recent shape-from-shading methods [46, 51]. However, the most practical next step is to scale our approach to room-scale scenes instead of isolated objects; e.g., on ScanNet data [7].

### Acknowledgments

We gratefully acknowledge Google's support of this project. This research is funded by a Google Tango grant, and supported by a Stanford Graduate Fellowship. We also gratefully acknowledge hardware donations from NVIDIA Corporation. We want to thank Ivan Dryanovski and Jürgen Sturm for their valuable feedback and help during this project, and Wenzel Jakob for the Mitsuba raytracer [15].

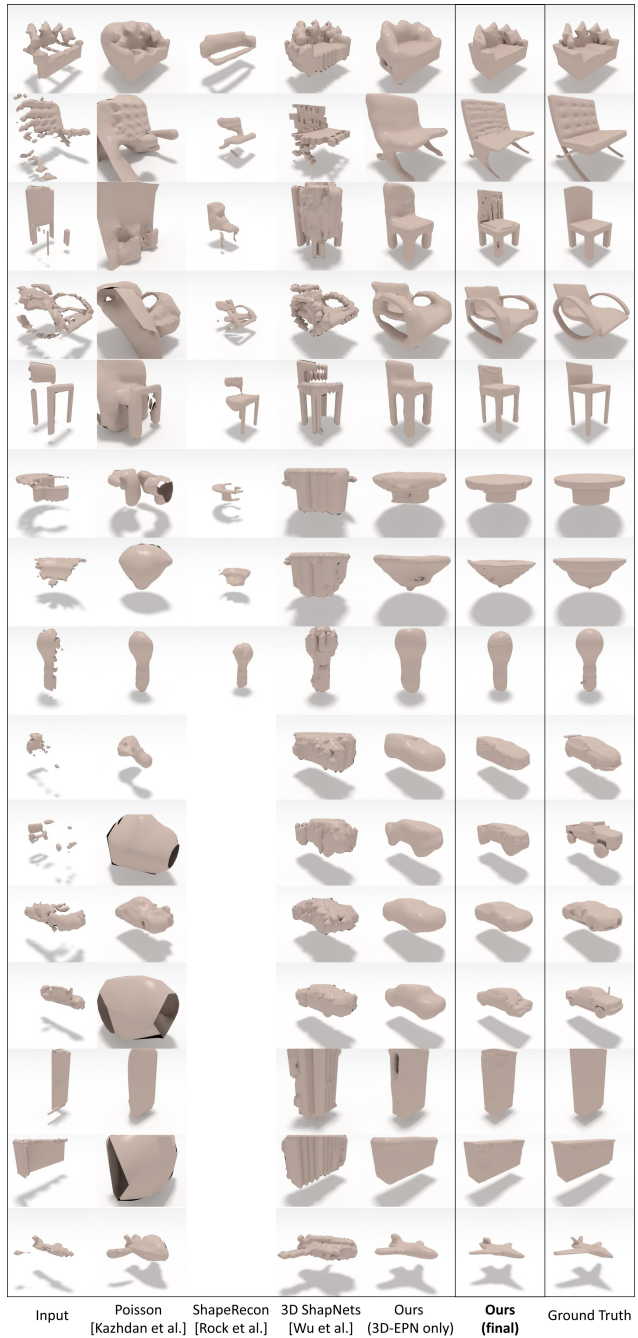

Figure 4: Qualitative evaluation on ShapeNet [2]. We show results on a variety of different scenes and compare against [16, 34, 48]. ShapeRecon is only trained on a subset of categories (top rows). We also show intermediate results where we only use the 3D-EPN w/o 3D shape synthesis. Input is visualized at $32^3$; however, for Kazhdan et al. [16] and Rock et al. [34], we use the $128^3$ input. We compare favorably, even only the 3D-EPN, but final shape synthesis increases the resolution and adds additional geometric detail.

# References

[1] J. Amanatides, A. Woo, et al. A fast voxel traversal algorithm for ray tracing. In *Eurographics*, volume 87, pages 3–10, 1987. 3, 4

[2] A. X. Chang, T. Funkhouser, L. Guibas, P. Hanrahan, Q. Huang, Z. Li, S. Savarese, M. Savva, S. Song, H. Su, J. Xiao, L. Yi, and F. Yu. ShapeNet: An Information-Rich 3D Model Repository. Technical Report arXiv:1512.03012 [cs.GR], Stanford University — Princeton University — Toyota Technological Institute at Chicago, 2015. 3, 4, 7, 8, 11

[3] J. Chen, D. Bautembach, and S. Izadi. Scalable real-time volumetric surface reconstruction. *ACM Transactions on Graphics (TOG)*, 32(4):113, 2013. 1

[4] S. Choi, Q.-Y. Zhou, and V. Koltun. Robust reconstruction of indoor scenes. In *2015 IEEE Conference on Computer Vision and Pattern Recognition (CVPR)*, pages 5556–5565. IEEE, 2015. 1

[5] S. Choi, Q.-Y. Zhou, S. Miller, and V. Koltun. A large dataset of object scans. *arXiv preprint arXiv:1602.02481*, 2016. 11

[6] B. Curless and M. Levoy. A volumetric method for building complex models from range images. In *Proceedings of the 23rd annual conference on Computer graphics and interactive techniques*, pages 303–312. ACM, 1996. 4

[7] A. Dai, A. X. Chang, M. Savva, M. Halber, T. Funkhouser, and M. Nießner. Scannet: Richly-annotated 3d reconstructions of indoor scenes. *arXiv preprint arXiv:1702.04405*, 2017. 8

[8] A. Dai, M. Nießner, M. Zollöfer, S. Izadi, and C. Theobalt. Bundlefusion: Real-time globally consistent 3d reconstruction using on-the-fly surface re-integration. *arXiv preprint arXiv:1604.01093*, 2016. 1

[9] A. Dai, C. R. Qi, and M. Nießner. Shape completion using 3d-encoder-predictor cnns and shape synthesis. In *Proceedings of the IEEE Conference on Computer Vision and Pattern Recognition*, pages –, 2017. 10

[10] A. Dosovitskiy, J. Tobias Springenberg, and T. Brox. Learning to generate chairs with convolutional neural networks. In *Proceedings of the IEEE Conference on Computer Vision and Pattern Recognition*, pages 1538–1546, 2015. 3

[11] M. Firman, O. Mac Aodha, S. Julier, and G. J. Brostow. Structured prediction of unobserved voxels from a single depth image. In *Proceedings of the IEEE Conference on Computer Vision and Pattern Recognition*, pages 5431–5440, 2016. 3

[12] I. Goodfellow, J. Pouget-Abadie, M. Mirza, B. Xu, D. Warde-Farley, S. Ozair, A. Courville, and Y. Bengio. Generative adversarial nets. In *Advances in Neural Information Processing Systems*, pages 2672–2680, 2014. 3, 8

[13] A. Hertzmann, C. E. Jacobs, N. Oliver, B. Curless, and D. H. Salesin. Image analogies. In *Proceedings of the 28th annual conference on Computer graphics and interactive techniques*, pages 327–340. ACM, 2001. 5

[14] P. Indyk and R. Motwani. Approximate nearest neighbors: towards removing the curse of dimensionality. In *Proceedings of the thirtieth annual ACM symposium on Theory of computing*, pages 604–613. ACM, 1998. 5

[15] W. Jakob. Mitsuba renderer, 2010. *URL: http://www.mitsuba-renderer.org*, 3, 2015. 8

[16] M. Kazhdan, M. Bolitho, and H. Hoppe. Poisson surface reconstruction. In *Proceedings of the fourth Eurographics symposium on Geometry processing*, volume 7, 2006. 1, 2, 7, 8

[17] M. Kazhdan and H. Hoppe. Screened poisson surface reconstruction. *ACM Transactions on Graphics (TOG)*, 32(3):29, 2013. 1, 2, 7

[18] Y. M. Kim, N. J. Mitra, D.-M. Yan, and L. Guibas. Acquiring 3d indoor environments with variability and repetition. *ACM Transactions on Graphics (TOG)*, 31(6):138, 2012. 2

[19] D. Kingma and J. Ba. Adam: A method for stochastic optimization. *arXiv preprint arXiv:1412.6980*, 2014. 5

[20] C. Li and M. Wand. Combining markov random fields and convolutional neural networks for image synthesis. *arXiv preprint arXiv:1601.04589*, 2016. 3

[21] C. Li and M. Wand. Precomputed real-time texture synthesis with markovian generative adversarial networks. *arXiv preprint arXiv:1604.04382*, 2016. 3

[22] D. Li, T. Shao, H. Wu, and K. Zhou. Shape completion from a single rgbd image. 2016. 2

[23] Y. Li, A. Dai, L. Guibas, and M. Nießner. Database-assisted object retrieval for real-time 3d reconstruction. In *Computer Graphics Forum*, volume 34, pages 435–446. Wiley Online Library, 2015. 2

[24] D. Maturana and S. Scherer. Voxnet: A 3d convolutional neural network for real-time object recognition. In *Intelligent Robots and Systems (IROS), 2015 IEEE/RSJ International Conference on*, pages 922–928. IEEE, 2015. 3

[25] N. J. Mitra, L. J. Guibas, and M. Pauly. Partial and approximate symmetry detection for 3d geometry. In *ACM Transactions on Graphics (TOG)*, volume 25, pages 560–568. ACM, 2006. 2

[26] L. Nan, K. Xie, and A. Sharf. A search-classify approach for cluttered indoor scene understanding. *ACM Transactions on Graphics (TOG)*, 31(6):137, 2012. 2

[27] A. Nealen, T. Igarashi, O. Sorkine, and M. Alexa. Laplacian mesh optimization. In *Proceedings of the 4th international conference on Computer graphics and interactive techniques in Australasia and Southeast Asia*, pages 381–389. ACM, 2006. 1, 2

[28] R. A. Newcombe, S. Izadi, O. Hilliges, D. Molyneaux, D. Kim, A. J. Davison, P. Kohi, J. Shotton, S. Hodges, and A. Fitzgibbon. Kinectfusion: Real-time dense surface mapping and tracking. In *Mixed and augmented reality (ISMAR), 2011 10th IEEE international symposium on*, pages 127–136. IEEE, 2011. 1

[29] D. T. Nguyen, B.-S. Hua, M.-K. Tran, Q.-H. Pham, and S.-K. Yeung. A field model for repairing 3d shapes. In *The IEEE Conference on Computer Vision and Pattern Recognition (CVPR)*, volume 5, 2016. 3

[30] M. Nießner, M. Zollhöfer, S. Izadi, and M. Stamminger. Real-time 3d reconstruction at scale using voxel hashing. *ACM Transactions on Graphics (TOG)*, 2013. 1, 11

[31] M. Pauly, N. J. Mitra, J. Giesen, M. H. Gross, and L. J. Guibas. Example-based 3d scan completion. In *Sym-*

*posium on Geometry Processing*, number EPFL-CONF-149337, pages 23–32, 2005. 2

[32] M. Pauly, N. J. Mitra, J. Wallner, H. Pottmann, and L. J. Guibas. Discovering structural regularity in 3d geometry. In *ACM transactions on graphics (TOG)*, volume 27, page 43. ACM, 2008. 2

[33] C. R. Qi, H. Su, M. Nießner, A. Dai, M. Yan, and L. Guibas. Volumetric and multi-view cnns for object classification on 3d data. In *Proc. Computer Vision and Pattern Recognition (CVPR), IEEE*, 2016. 3, 4, 5, 7, 10

[34] J. Rock, T. Gupta, J. Thorsen, J. Gwak, D. Shin, and D. Hoiem. Completing 3d object shape from one depth image. In *Proceedings of the IEEE Conference on Computer Vision and Pattern Recognition*, pages 2484–2493, 2015. 2, 7, 8

[35] O. Ronneberger, P. Fischer, and T. Brox. U-net: Convolutional networks for biomedical image segmentation. In *International Conference on Medical Image Computing and Computer-Assisted Intervention*, pages 234–241. Springer, 2015. 4

[36] T. Shao, W. Xu, K. Zhou, J. Wang, D. Li, and B. Guo. An interactive approach to semantic modeling of indoor scenes with an rgbd camera. *ACM Transactions on Graphics (TOG)*, 31(6):136, 2012. 2

[37] A. Sharma, O. Grau, and M. Fritz. Vconv-dae: Deep volumetric shape learning without object labels. *arXiv preprint arXiv:1604.03755*, 2016. 3

[38] Y. Shi, P. Long, K. Xu, H. Huang, and Y. Xiong. Data-driven contextual modeling for 3d scene understanding. *Computers & Graphics*, 55:55–67, 2016. 2

[39] I. Sipiran, R. Gregor, and T. Schreck. Approximate symmetry detection in partial 3d meshes. In *Computer Graphics Forum*, volume 33, pages 131–140. Wiley Online Library, 2014. 2

[40] S. Song and J. Xiao. Deep sliding shapes for amodal 3d object detection in rgb-d images. *arXiv preprint arXiv:1511.02300*, 2015. 3

[41] O. Sorkine and D. Cohen-Or. Least-squares meshes. In *Shape Modeling Applications, 2004. Proceedings*, pages 191–199. IEEE, 2004. 1, 2

[42] P. Speciale, M. R. Oswald, A. Cohen, and M. Pollefeys. A symmetry prior for convex variational 3d reconstruction. In *European Conference on Computer Vision*, pages 313–328. Springer, 2016. 2

[43] M. Sung, V. G. Kim, R. Angst, and L. Guibas. Data-driven structural priors for shape completion. *ACM Transactions on Graphics (TOG)*, 34(6):175, 2015. 2, 12

[44] S. Thrun and B. Wegbreit. Shape from symmetry. In *Tenth IEEE International Conference on Computer Vision (ICCV'05) Volume 1*, volume 2, pages 1824–1831. IEEE, 2005. 2

[45] S. Whelan, S. Leutenegger, R. F. Salas-Moreno, B. Glocker, and A. J. Davison. Elasticfusion: Dense slam without a pose graph. *Proc. Robotics: Science and Systems, Rome, Italy*, 2015. 1

[46] C. Wu, M. Zollhöfer, M. Nießner, M. Stamminger, S. Izadi, and C. Theobalt. Real-time shading-based refinement for consumer depth cameras. *ACM Transactions on Graphics (TOG)*, 33(6), 2014. 8

[47] J. Wu, C. Zhang, T. Xue, W. T. Freeman, and J. B. Tenenbaum. Learning a probabilistic latent space of object shapes via 3d generative-adversarial modeling. *arXiv preprint arXiv:1610.07584*, 2016. 3

[48] Z. Wu, S. Song, A. Khosla, F. Yu, L. Zhang, X. Tang, and J. Xiao. 3d shapenets: A deep representation for volumetric shapes. In *Proceedings of the IEEE Conference on Computer Vision and Pattern Recognition*, pages 1912–1920, 2015. 3, 7, 8, 12

[49] A. Zeng, S. Song, M. Nießner, M. Fisher, and J. Xiao. 3dmatch: Learning the matching of local 3d geometry in range scans. *arXiv preprint arXiv:1603.08182*, 2016. 3

[50] W. Zhao, S. Gao, and H. Lin. A robust hole-filling algorithm for triangular mesh. *The Visual Computer*, 23(12):987–997, 2007. 1, 2

[51] M. Zollhöfer, A. Dai, M. Innmann, C. Wu, M. Stamminger, C. Theobalt, and M. Nießner. Shading-based refinement on volumetric signed distance functions. *ACM Transactions on Graphics (TOG)*, 2015. 8

# Appendix

In this appendix, we provide additional evaluation and results of our shape completion method "Shape Completion using 3D-Encoder-Predictor CNNs and Shape Synthesis" [9].

## A. Additional Results on Synthetic Scans

Tab. 3 shows a quantitative evaluation of our network on a test set of input partial scans with varying trajectory sizes ($\geq 1$ camera views). Our 3D-EPN with skip connections and class vector performs best, informing the best shape synthesis results.

| Method | $\ell_1$-Err ($32^3$) | $\ell_1$-Err ($128^3$) |
|---|---|---|
| Ours (3D-EPN + synth) | 0.382 | 1.94 |
| Ours (3D-EPN-class + synth) | 0.376 | 1.93 |
| Ours (3D-EPN-unet + synth) | 0.310 | 1.82 |
| **Ours (final)** 3D-EPN-unet-class + synth | **0.309** | **1.80** |

Table 3: Quantitative shape completion results on synthetic ground truth data for input partial scans with varying trajectory sizes. We measure the $\ell_1$ error of the unknown regions against the ground truth distance field (in voxel space, up to truncation distance of 2.5 voxels).

## B. Results on Real-world Range Scans

In Fig. 8, we show example shape completions on real-world range scans. The test scans are part of the RGB-D test set of the work of Qi et al. [33], and have been captured with a PrimeSense sensor. The dataset includes reconstructions and frame alignment obtained through VoxelHashing [30] as well as mesh objects which have been manually segmented from the surrounding environment. For the purpose of testing our mesh completion method, we only use the first depth frame as input (left column of Fig. 8). We use our 3D-EPN trained as described on purely synthetic data from ShapeNet [2]. As we can see, our method is able to produce faithful completion results even for highly partial input data. Although the results are compelling for both the intermediate 3D-EPN predictions, as well our final output, the completion quality looks visually slightly worse than the test results on synthetic data. We attribute this to the fact that the real-world sensor characteristics of the PrimeSense are different from the synthetically-generated training data used to train our model. We believe a better noise model, reflecting the PrimeSense range data, could alleviate this problem (at the moment we don't simulate sensor noise). Another option would be to generate training data from real-world input, captured with careful scanning and complete scanning patterns; e.g., using the dataset captured by Choi et al. [5]. However, we did not further explore this direction in the context of the paper, as our goal was to learn the completions from actual ground truth input. In addition to 3D-EPN predictions and our final results, we show the intermediate shape retrieval results. These models are similar; however, they differ significantly from the partial input with respect to global geometric structure. Our final results thus combine the advantages of both the global structure inferred by our 3D-EPN, as well as the local detail obtained through the shape synthesis optimization process.

## C. Evaluation on Volumetric Representation

In Table 4, we evaluate the effect of different volumetric surface representations. There are two major characteristics of the representation which affect the 3D-EPN performance. First, a smooth function provides better performance (and super-resolution encoding) than a discrete representation; this is realized with signed and unsigned distance fields. Second, explicitly storing known-free space encodes information in addition to the voxels on the surface; this is realized with a ternary grid and the sign channel in the signed distance field. The signed distance field representation combines both advantages.

## D. Single Class vs Multi-Class Training

Table 5 evaluates different training options for performance over multiple object categories. We aim to answer the question whether we benefit from training a separate network for each class separately (first column). Table 5 compares the results of training separate networks for each class with a single network trained over all classes (with and without class information). Our networks trained over all classes combined performs better than training over each individual class, as there is significantly more training data, and the network leveraging class predictions performs the best.

| Surface Rep. | $\ell_1$-Error ($32^3$) | $\ell_2$-Error ($32^3$) |
|---|---|---|
| Binary Grid | 0.653 | 1.160 |
| Ternary Grid | 0.567 | 0.871 |
| Distance Field | 0.417 | 0.483 |
| Signed Distance Field | **0.379** | **0.380** |

Table 4: Quantitative evaluation of the surface representation used by our 3D-EPN. In our final results, we use a signed distance field input; it encodes the ternary state of known-free space, surface voxels, and unknown space, and is a smooth function. It provides the lowest error compared to alternative volumetric representations.

| Category (# train models) | Separate EPN-unets (known class) $\ell_1$-Error | EPN-unet w/o Class $\ell_1$-Error | EPN-unet /w Class **Ours Final** $\ell_1$-Error |
|---|---|---|---|
| Chairs (5K) | 0.477 | **0.409** | 0.418 |
| Tables (5K) | 0.423 | **0.368** | 0.377 |
| Sofas (2.6K) | 0.478 | 0.421 | **0.392** |
| Lamps (1.8K) | 0.450 | 0.398 | **0.388** |
| Planes (3.3K) | 0.440 | **0.418** | 0.421 |
| Cars (5K) | 0.271 | 0.266 | **0.259** |
| Dressers (1.3K) | 0.453 | 0.387 | **0.381** |
| Boats (1.6K) | 0.380 | 0.364 | **0.356** |
| Total (25.7K) | 0.422 | 0.379 | **0.374** |

Table 5: Quantitative evaluations of $32^3$ 3D-EPNs; from left to right: separate networks have been trained for each class independently (at test time, the ground truth class is used to select the class network); a single network is used for all classes, but no class vector is used; our final result uses a single network trained across all classes and we input a probability class vector into the latent space of the 3D-EPN.

## E. Evaluation on Different Degrees of Incompleteness

Fig. 5 shows an evaluation and comparisons against 3D ShapeNets [2] on different test datasets with varying degrees of partialness. Even for highly partial input, our method achieves relatively low completion errors. Compared to previous work, the error rate of our method is relatively stable with respect to the degree of missing data.

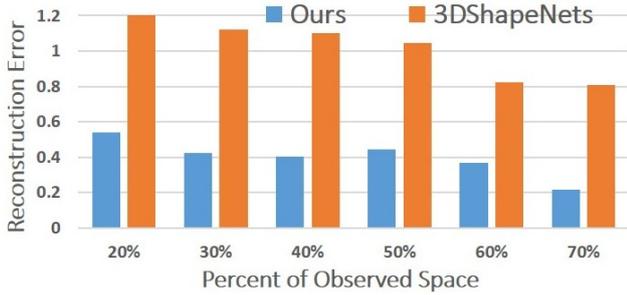

Figure 5: Quantitative evaluation of shape completion using our 3D-EPN and 3D ShapeNets [48] on different degrees of partial input. For this task, we generate several test sets with partial observed surfaces ranging from 20% to 70%. Even for very partial input, we obtain relatively low reconstruction errors, whereas 3D ShapeNets becomes more unstable.

## F. Comparison against Sung et al. [43]

In Tab. 6 and Fig. 6, we compare against the method by Sung et al. [43] using the dataset published along with their method. Note that their approach operates on a point cloud representation for both in and output. In order to provide a fair comparison, we apply a distance transform of the predicted points and measure the $\ell_1$ error on a $32^3$ voxel grid.

| Class (#models) | $\ell_1$-Error ($32^3$) | |
|---|---|---|
| | Sung et. al [43] | Ours |
| assembly_airplanes (58) | 0.56 | **0.50** |
| assembly_chairs (64) | 0.73 | **0.51** |
| coseg_chairs (287) | 0.72 | **0.57** |
| shapenet_tables (37) | 0.82 | **0.45** |
| Total (446) | 0.71 | **0.54** |

Table 6: Quantitative shape completion results on the dataset of Sung et. al [43]. We measure the $\ell_1$ error of the unknown regions against the ground truth distance field (in voxel space, up to truncation distance of 3 voxels).

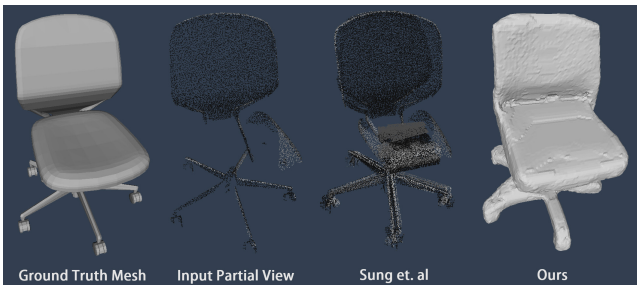

Figure 6: Qualitative comparison against Sung et. al [43]. Note that the missing chair seat and front of chair back introduce difficulties for inferring structure, whereas our method is able to more faithfully infer the global structure.

## G. Shape Embeddings

Fig. 7 shows a t-SNE visualization of the latent vectors in our 3D-EPN trained for shape completion. For a set of test input partial scans, we extract their latent vectors (the 512-dimensional vector after the first fully-connected layer and before up-convolution) and then use t-SNE to reduce their dimension to 2 as $(x, y)$ coordinates. Images of the partial scans are displayed according to these coordinates. Shapes with similar geometry tend to lie near each other, although they have varying degrees of occlusion.

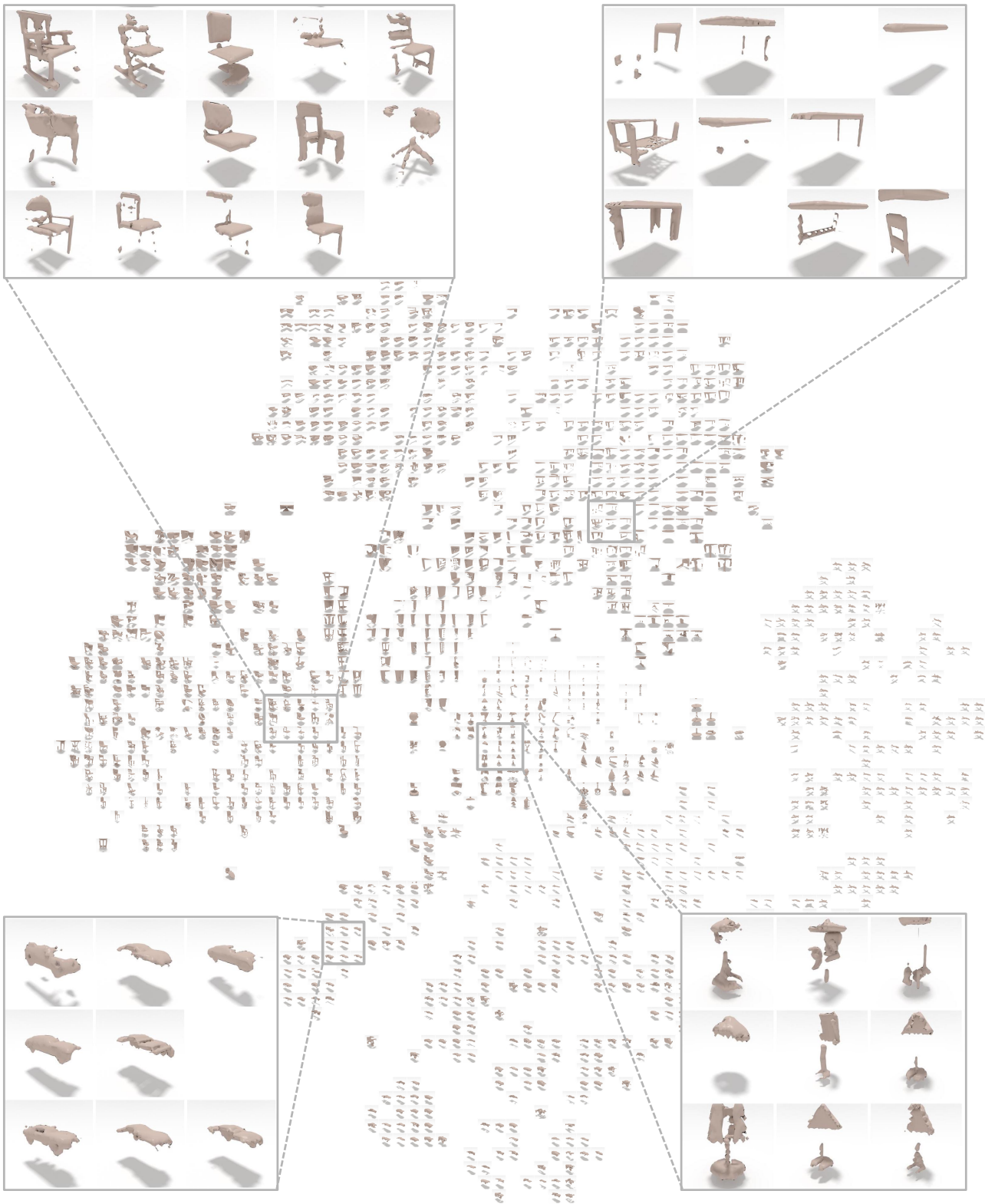

Figure 7: t-SNE visualization of the latent vectors in our 3D-EPN trained for shape completion. The rendered images show input partial scans. Four zoom-ins are shown for regions of chairs (top left), tables (top right), cars (bottom left) and lamps (bottom right).

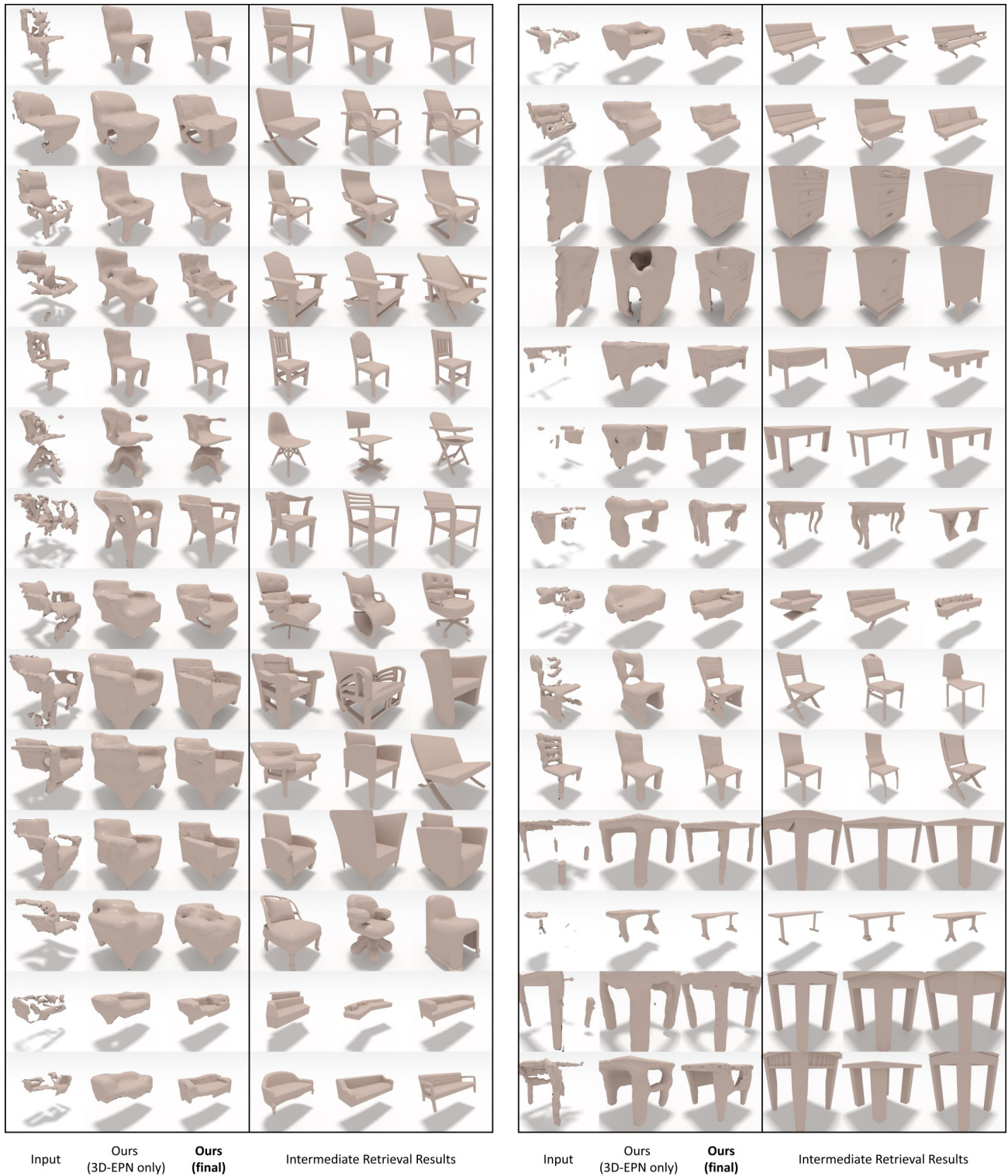

Figure 8: Example shape completions from our method on real-world range scans from commodity sensors (here, a PrimeSense is used). We visualize partial input, 3D-EPN predictions, and our final results. In addition, we show the retrieved shapes as intermediate results on the right. Note that although the retrieved models look clean, they are inherently different from the input with respect to global structure.